\documentclass[letterpaper, 10 pt, conference]{ieeeconf}  

\IEEEoverridecommandlockouts                              

\usepackage{booktabs}       
\usepackage{amsfonts}       
\usepackage{nicefrac}       
\usepackage{pifont}
\usepackage{xcolor}         
\usepackage{multirow}
\usepackage{microtype}      
\usepackage{graphicx} 
\usepackage{amsmath}
\usepackage{amssymb}
\usepackage{wrapfig}
\usepackage{colortbl}
\usepackage{url}
\usepackage[colorlinks,citecolor=ForestGreen]{hyperref}

\usepackage{cuted}   
\usepackage{capt-of} 

\newlength{\smallimage}
        \definecolor{rel}{rgb}{.1,.6,.2}
        \definecolor{nrl}{rgb}{1,1,1}
        \definecolor{qim}{rgb}{1,1,1}

\definecolor{lightgray}{gray}{0.93}

\def\be{\begin{equation}}
\def\ee{\end{equation}}
\def\bea{\begin{eqnarray}}
\def\eea{\end{eqnarray}}
\def\ben{\begin{eqnarray*}}
\def\een{\end{eqnarray*}}

\def\bi{\begin{itemize}}
\def\ei{\end{itemize}}

\newcommand{\btab}[1]{\begin{tabular}{#1}}
\newcommand{\etab}{\end{tabular}}
\newcommand{\ba}[1]{\begin{array}{#1}}
\newcommand{\ea}{\end{array}}

\def\<{\langle}
\def\>{\rangle}

\newcommand{\myparagraph}[1]{\vspace{0.1cm}\noindent\textit{#1.}}



\definecolor{DarkGreen}{rgb}{0.5, 0.9, 0.5}

\newcommand{\duster}[0]{DUSt3R}
\newcommand{\master}[0]{MASt3R}

\newcommand{\muster}[0]{MUSt3R}
\newcommand{\slider}[0]{S-MUSt3R}

\newcommand{\vggt}[0]{VGGT}
\newcommand{\vggtlong}[0]{VGGT-Long}

\usepackage{xstring}
\usepackage{hhline}
\usepackage{siunitx}
\usepackage{tikz}
\usepackage{textcomp}
\usepackage{array,multirow,graphicx}
\usepackage{stmaryrd}
\usepackage{multicol}
\usepackage{hyperref}
\usepackage{xspace}
\usepackage{cleveref}
\usepackage{wrapfig}

\newcommand{\SEtwo}[1]{\ensuremath{\mathrm{SE}(2)}\xspace}
\newcommand{\SE}[1]{\ensuremath{\mathrm{SE}(#1)}\xspace}

\newcommand{\SLfour}{\ensuremath{\mathrm{SL}(4)}\xspace}

\newcommand{\SEthree}[1]{\ensuremath{\mathrm{SE}(3)}\xspace}

\definecolor{myemerald}{rgb}{0.753, 0.898, 0.804}
\definecolor{mylightgreen}{rgb}{0.894, 0.933, 0.745}
\definecolor{myyellow}{rgb}{0.996, 0.972, 0.780}

\newcommand{\firstc}{\cellcolor{myemerald!100}}
\newcommand{\secondc}{\cellcolor{mylightgreen!100}}
\newcommand{\thirdc}{\cellcolor{myyellow!100}}

\newcommand{\cg}{\color{gray!70}}

\definecolor{Gray}{gray}{0.85}
\definecolor{GrayBorder}{gray}{0.65}
\definecolor{green_cylinder}{rgb}{0.0,0.40,0.0}
\definecolor{blue_cylinder}{rgb}{0.02,0.05,0.75}
\definecolor{way_point}{rgb}{0.56,0.0,1.0}
\newcolumntype{a}{>{\columncolor{GrayBorder}}c}

\title{\LARGE \bf
S-MUSt3R: Sliding Multi-view 3D Reconstruction
}

\author{Leonid Antsfeld, Boris Chidlovskii, Yohann Cabon, Vincent Leroy, Jerome Revaud} 

\begin{document}
\maketitle

\vspace{-1em}
\begin{strip}
  \centering
  \includegraphics[width=0.9\textwidth]{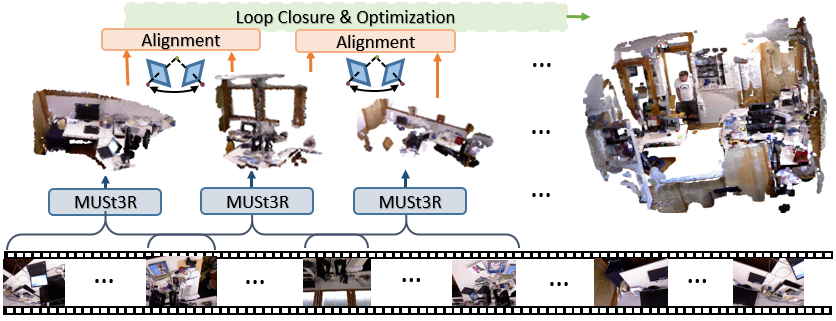}%
  \vspace{-0.5em}
  \captionof{figure}{\slider{} partitions the long image sequence in segments, applies \muster{} model, then aligns the segments and corrects the accumulated drift with lightweight loop closure and optimization.}
  \vspace{0.5em}
\end{strip}

\begin{abstract}
The recent paradigm shift in 3D vision led to the rise of foundation models with remarkable capabilities in 3D perception from uncalibrated images. However, extending these models to large-scale RGB stream 3D reconstruction remains challenging due to memory limitations. This work proposes \slider{}, a simple and efficient pipeline that extends the limits of foundation models for monocular 3D reconstruction. 
Our approach addresses the scalability bottleneck of foundation models through a simple strategy of sequence segmentation followed by segment alignment and lightweight loop closure optimization. Without model retraining, we benefit from remarkable 3D reconstruction capacities of \muster{} model and achieve trajectory and reconstruction performance comparable to traditional methods with more complex architecture. We evaluate \slider{} on TUM, 7-Scenes and proprietary robot navigation datasets and show that \slider{} runs successfully on long RGB sequences 
and produces accurate and consistent 3D reconstruction. Our results highlight the potential of leveraging the \muster{} model for scalable monocular 3D scene in real-world settings, with an important advantage of making predictions directly in the metric space.
\end{abstract}

\section{Introduction}
\label{sec:intro}

Perceiving three-dimensional environments from monocular RGB streams is a fundamental capability for autonomous systems, yet current methodologies remain limited when applied to long uncalibrated sequences. 
Existing approaches \cite{gigaslam,dvso,dfvo} have demonstrated progress on large-scale monocular scenes, but they frequently rely on multi-component pipelines or presuppose accurate camera intrinsics. Alternative solutions integrate auxiliary sensing modalities such as LiDAR~\cite{zhang2014loam}, IMU~\cite{zhang2015visual}, or stereo vision~\cite{cvivsic2022soft2}, thereby circumventing the core problem of monocular reconstruction. However, scalable and calibration-free 3D reconstruction from monocular RGB alone remains an open challenge, and addressing it is critical for a broad range of robotic and embodied agent navigation tasks in diverse real-world environments.

A paradigm shift comes from the 3D vision domain which has recently witnessed the rise of feed-forward foundation models based on the Transformer architecture~\cite{vaswani2017attention}. The seminal work \duster{}~\cite{wang2024dust3r} followed by \master{}~\cite{master}, \master-SfM~\cite{mast3r-sfm}, CUT3R~\cite{cut3r} and FASt3R~\cite{faster}, aim to replace multi-component SfM and SLAM pipelines with a single and unified deep learning network. 
Trained on massive datasets with backpropagation of errors through the entire pipeline integrating 3D scene representation, camera pose and intrinsic parameter estimation, these models enable robust 3D reconstruction from one or a few uncalibrated RGB images. They predict pointmaps, depths and camera poses in a single forward pass creating powerful foundation models for end-to-end 3D scene understanding. 

The most advanced models, such as VGGT~\cite{vggt}  and \muster{}~\cite{cabon25muster}, achieve state-of-the-art local reconstructions, but their scalability is constrained by the heavy computational and memory demands of transformer-based architectures. While techniques like Flash-Attention~\cite{dao2023flashattention2} improve efficiency, GPU memory consumption remains high, restricting these models to relatively short sequences. Even high-end GPUs process only a few hundred frames before memory limits are reached, making large-scale reconstructions unfeasible. 

To overcome this limitation, recent efforts integrate foundation models into larger systems. One example is \master-SLAM~\cite{mast3r-slam} which combines \master{} with a complex backend of pose graph optimization and bundle adjustment. In contrast, \vggt-Long~\cite{vggt-long} adopts a {\it minimalist} approach, demonstrating that foundation models themselves can serve as a powerful engine for large-scale 3D perception with limited additional overhead. 

In a similar spirit, we propose \slider{}, a sliding-window extension of \muster{}~\cite{cabon25muster} for long sequences. Our framework partitions input videos into overlapping segments, reconstructs each independently, then aligns and stitches them with a lightweight loop closure module. This approach requires no model retraining, it exploits the strong local accuracy of \muster{}, while it also addresses drift and scalability issues without requiring a sophisticated graph-based optimization backend. The result is a globally consistent 3D reconstruction system that remains simple, efficient, and directly applicable to downstream tasks such as robot navigation \cite{ObjectNavRevisited}. Our choice of \muster{} as the foundation model is guided by its capacity to make predictions directly in the metric space.

Our work makes the following key contributions:
\begin{enumerate}
    \item We extend \muster{} foundation model to large-scale scenes, without requiring camera calibration. We adopt a segment-process-stitch approach that mitigates memory constraints on long sequences.
    \item We propose multiple improvements which address drift accumulation with lightweight alignment and loop closure, showing that \muster{} can scale without a complex backend.
    \item We provide extensive quantitative comparisons to state-of-the-art uncalibrated methods and show that \slider{} is on par with \master-SLAM and \vggt-SLAM which deploy sophisticated graph-based backends, and outperforms by a large margin the stitching-based competitor method \vggt-Long~\cite{vggt-long}.
\end{enumerate}

\section{Related Work}
\label{sec:related}

\myparagraph{Classical scene reconstruction}
Traditional scene reconstruction methods rely on geometric features to estimate camera poses and reconstruct 3D scenes from multi-view images~\cite{Cramariuc22ral-Maplab,Furukawa10cvpr,Rosinol20icra-Kimera}. Using multi-view geometry~\cite{Koenderink91} and bundle adjustment~\cite{Pan24eccv-GLOMAP,Schonberger16cvpr-SfMRevisited}, these pipelines extract sparse features, match them, and optimize for \SE{3} transformations.  
Learning-based methods depart from handcrafted SfM~\cite{disco2013pami,hartleymultiviewgeometry,jiang13,colmapsfm} and aim to directly predict 3D geometry and camera parameters from one or more RGB images. Neural networks learn strong 3D priors, either object-centric~\cite{zero-1-to-3,shapefromsingleimg23} or scene-level~\cite{smith24flowmap,deepv2d,WangCVPR24VisualGeometryGroundedDeepSfM,deeptam}, often via differentiable SfM trained end-to-end.  

\myparagraph{Transformer-based feed-forward reconstruction}
The seminal work~\duster{}~\cite{wang2024dust3r} introduced feed-forward dense stereo reconstruction from image pairs, producing point maps without requiring known intrinsics. Its successor~\master~\cite{master} improved pairwise matching with learned descriptors and was later extended to dynamic scenes~\cite{zhang24monster} and multi-image optimization~\cite{mast3r-sfm}.  
To move beyond pairwise processing, several works proposed memory or recurrent mechanisms. Spann3R~\cite{WangX24Spanner3DReconstructionWithSpatialMemory} employed a spatial memory to track observations in sequential pairs. 
CUT3R~\cite{cut3r} used recurrent state modules for incremental multi-view reconstruction, while Pow3R~\cite{Jang25arxiv-Pow3R} improved robustness by incorporating optional intrinsics, poses, or depth.  

\muster{}~\cite{cabon25muster} generalized \duster{} framework to  $N$-view settings with a multi-layer memory for representing scenes as pointmaps, enabling direct image-to-image correspondences, Fast3R~\cite{faster} replaced pairwise attention with all-to-all attention, \vggt~\cite{vggt} further extended Transformers to output cameras, depths, pointmaps, and tracks, all scaling to hundreds of views in one pass.  

\myparagraph{SLAM on uncalibrated images}
Traditional SLAM systems~\cite{orbslam3,orbslam,orbslam2} rely on geometric optimization, while learning-based approaches~\cite{gradslam,Teed21nips-DROID-SLAM} integrate differentiable components. To exploit 3D foundation models, \master-SLAM~\cite{mast3r-slam} built real-time monocular pipelines around~\master{}~\cite{master}, including efficient transform estimation and loop closure. \vggt-SLAM~\cite{maggio2025vggtslamdensergbslam} instead incrementally aligned \vggt{} submaps through $\SLfour$ optimization, overcoming the limitations of 
similarity-based alignment. 

The closest to ours is \vggt-Long~\cite{vggt-long} which deploys \vggt~\cite{vggt} as the foundation model for processing sequence segments and SIM(3) groups to align them in one global 3D representation. However, \vggt{} output is non-metric, and the robot needs additional cues to resolve scale, combine it with stereo, LiDAR or wheel odometry to recover metric scale. Unlike \vggt-Long, \slider{} uses \muster{} as the foundation model to process individual segments, it takes advantage from \muster{} producing scene geometry in absolute physical units that can be used on the real robots. The robot can directly use the predictions to estimate traversability, collision avoidance and in path planning. 

\section{Architecture}
\label{sec:architecture}

\subsection{\muster{}}
\label{sec:muster}
\muster{}~\cite{cabon25muster} extends pair-wise \duster{} to an arbitrary number of images and maps them in 3D pointmaps in a first frame's coordinate system. It uses a multi-layered memory, 
which contains patches of previously seen images. 
To control the memory size growing linearly with the number of images, 
\muster{} applies a special strategy to carefully select memory tokens using the image discovery rate; it leverages a running memory and updates 3D scene of current observations on-the-fly. . 
For an input image $I$ of size ${H\times W}$, \muster{} outputs its pointmap ${\bf X} \in R^{3 \times H\times W}$, confidence map ${\bf C} \in R^{H\times W}$ and depth map ${\bf d} \in R^{H\times W}$.  
\muster{} is able to process hundreds of images, but hits the memory limits on longer sequences. In the next section we complement \muster{} with additional steps of segmenting long input sequences and stitching local 3D pointmaps in the global one. We propose multiple modifications to the aligning, stitching and loop closure in order to ensure a robust 3D reconstruction.

\subsection{\slider}
\label{sec:slider}

\slider{} is a sliding version of \muster{} running over long monocular image sequences. First, it splits the sequence into overlapping segments; second, \muster{} processes segments one by one; third, it aligns the segments to express in the first frame's reference, correct the final representation by detecting segment-wise loops, building a pose graph where segments are nodes and the edges are constrained by alignments, followed by pose graph optimization. Thus we benefit from local reconstruction of segments by \muster{} while ensuring global accuracy when fast and efficient collecting segments in the full dense scene pointmap.

An input sequence of $N$ images, $\{I_i\}_{i=1}^N$ is first partitioned into a number of overlapping segments. 
The generated segments all have the same length $l$ and the overlap size $p$. The first segment $\mathcal{S}_1$ includes frames from 1 to $l$, the second segment $\mathcal{S}_2$ includes frames from $l-p+1$ to $2*l-p$, and so on. 

\subsection{Confidence weighted by depth difference}
\label{sec:confidence}
\muster{} trains the model using a confidence-aware loss and predicts a confidence score for each pixel in the images.
The segment alignment is sensible to 3D outliers and accurate confidence maps are critical for filtering the outliers out. We therefore take advantage from segment overlaps as an additional source of information for confidence estimation. The same image gets different context in adjacent segments, and \muster{} model often output different depth estimation for the same image. Any inconsistency in depth estimation can undermine the accurate segment alignment.

To align 3D pointmaps of two overlapping segments we use both confidence and depth maps; we trust points with higher confidence and down-weight points with inconsistent depth. 
Given confidence values $c_{ip}, c_{jp}$ and depth values $d_{ip}, d_{jp}$ for pixel $p$ of the image $I$ present in overlapping segments ${\cal S}_i$ and ${\cal S}_j$, we modulate the confidences by penalizing the depth disagreement with weight $w$. This weight  
\begin{equation}
w = \frac{c_{ip} \cdot c_{jp}}{1 + \lvert d_{ip} - d_{jp} \rvert}
\label{eq:weight}
\end{equation}
is used to update the confidence values $c'_{ip}= w \cdot c_{ip}$ and $c'_{jp} = w \cdot c_{jp}$. With some abuse of notation, in the following we denote the updated confidence maps as ${\bf C}_i, {\bf C}_j$.
 
\subsection{Transform groups}
\label{sec:transform}
Stitching local pointmaps in the global one relies on the accurate estimation of all segment alignments in order to express all 3D point coordinates in the first frame's reference. It is common to estimate transforms using SIM(3) Lie group~\cite{vggt-long,eth3d_slam,Teed21nips-DROID-SLAM} for which there exist close form and fast iterative solutions. However, the rigid transformations defined by SIM(3) group assume that input images are calibrated. In the case of uncalibrated images, it requires  relaxation~\cite{maggio2025vggtslamdensergbslam} as the rigid transformation can not cope with different distortions in the output of multi-view methods, like stretching and angular deformations of the scene geometry. Recent works propose to replace SIM(3) group with Affine(3)~\cite{yu2025relativeposeestimationaffine} or SL(4) groups~\cite{maggio2025vggtslamdensergbslam} to cope with such distortions. 

In the following we consider the alignment between two segments as belonging to transform group $\mathcal{T}$, where $\mathcal{T}$ is one of three Lie groups of increasing complexity, SIM(3), Affine(3) or SL(4). Conventional SIM(3) group includes rotations, translations and uniform scaling. 
Affine(3) group includes in addition non-uniform scaling and shearing. Then, SL(4) group includes translations, scaling and projective warping. We note however that a higher expressive power comes with a higher compute cost.
In Sec.~\ref{sec:exp}, we report the evaluation results and show that SIM(3) represents a globally best performance-speed compromise for our pipeline.

\begin{figure}[t]
    \centering
    \includegraphics[width=0.8\linewidth]{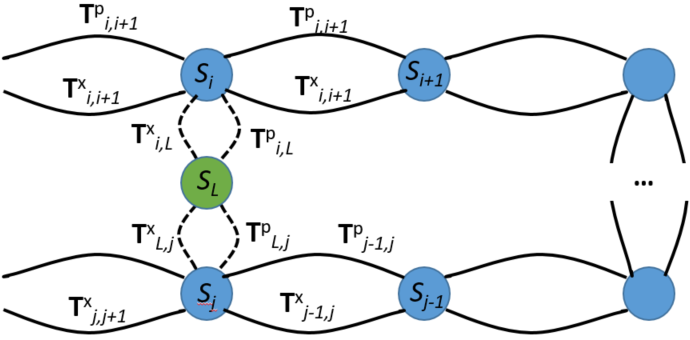}
    \caption{Example of a pose graph where segments are nodes (nodes in \textcolor{blue}{blue} for sequence segments and an extra node in \textcolor{green}{green} for loop closure) 
    and edges are constrained by pose-driven and pointmap-driven alignments.}
    \label{fig:graph}
\end{figure}

\subsection{Segment Aligning with Confidence and Depth}
\label{sec:align}
\muster{} processes segments independently; given an segment $\mathcal{S}_k$, it outputs a 3D pointmap ${\bf X}_k$, confidence map ${\bf C}_k$, per-frame depth estimation ${\bf d}_k$, completed with segment-wise consistent camera poses ${\bf p}_k$. We leverage \muster{}'s confidence as well the frame-based depth estimation to robustly align the overlapping segments. 

For two adjacent segments, $\mathcal{S}_k$ and $\mathcal{S}_{k+1}$, we identify a set of 3D point correspondences $({\bf X}_k^i, {\bf X}_{k+1}^i)$ and confidences $({\bf C}_k^i, {\bf C}_{k+1}^i)$ within their overlapping region. To robustly estimate the transformation ${\bf T}_{k,k+1} \in \mathcal{T}$ that aligns $\mathcal{S}_{k+1}$ to $\mathcal{S}_k$, we deploy, in the case of SIM(3) and Affine(3) transforms, the Iteratively Reweighted Least Squares (IRLS) optimization \cite{vggt-long}. Its objective is to minimize the following robust cost function
\begin{equation}
{\bf T}_{k,k+1}^{x} = \arg\min_{{\bf T} \in \mathcal{T}} \sum_i \rho \left( \| {\bf X}_k^i - {\bf T} {\bf X}_{k+1}^i \|_2 \right),
\label{eq:align}
\end{equation} 
where $\rho(\cdot)$ is the Huber loss function which down-weights the influence of outliers. The IRLS procedure solves this non-linear problem by iteratively minimizing a weighted sum of squared errors. We refer to \cite{vggt-long} for more detail.

In addition to aligning adjacent segments by 3D correspondences, we propose to align segments $\mathcal{S}_k$ and $\mathcal{S}_{k+1}$ by another \muster{}'s output, namely the camera poses $({\bf p}_k^i, {\bf p}_{k+1}^i)$, within the overlapping region. 
In Eq.(\ref{eq:align}), we replace the pointmap ${\bf X}_i^j$ by the set of estimated camera poses ${\bf p}_i^j$. Due to a much smaller size, the inference of the alignment from camera poses is negligible w.r.t. pointmaps, 
\begin{equation}
{\bf T}_{k,k+1}^{p} = \arg\min_{{\bf T} \in \mathcal{T}} \sum_i \rho \left( \| {\bf P}_k^i - {\bf T} {\bf P}_{k+1}^i \|_2 \right).
\label{eq:align_pose}
\end{equation} 
Therefore, for two adjacent segments $S_k$ and $S_{k+1}$ we obtain two transform estimations, ${\bf T}_{k,k+1}^{x}$ and ${\bf T}_{k,k+1}^{p}$, inferred from pointmaps and camera poses, respectively. In the pose graph, they form two edges connecting nodes $\mathcal{S}_k$ and $\mathcal{S}_{k+1}$ (blue nodes in Fig.~\ref{fig:graph}). 

Unlike SIM(3) and Affine(3) groups, aligning segments with transforms from SL(4) group requires estimating a relative homography matrix between the segments; we use h-solver from \vggt-SLAM~\cite{maggio2025vggtslamdensergbslam}.

\subsection{Loop Detection and Global Optimization} 
\label{sec:loop}
Long sequences result in the drift accumulation. We remove the drift by detecting and closing loops across the entire sequence. This process involves finding visual content shared by non-adjacent segments and robustly estimating the transform $\bf T \in \mathcal{T}$ between them. 

First, we reuse output of \muster{}'s encoder which generates patch features for any image $I$ in the sequence. They are average-pooled to obtain a compact global feature vector $\mathbf{f}$ which captures the scene geometry in the image. 
These global image descriptors are used to identify potential loop closure candidates. We create and maintain KDTree() structure $D$ of image descriptors; and for each descriptor, we perform an efficient nearest neighbor search in $D$ to find other images with high similarity. A pair of distant images $(I, I'), I \in \mathcal{S}_i, I' \in \mathcal{S}_j$, $|i - j| > 2$ forms a potential loop closure if their similarity score exceeds a threshold $\sigma_{sim}$. 
Two distant segments with at least $k_{min}=3$ loop closure candidates form a loop.

For validated loop pairs $(I,I')$, we generate an additional reconstruction of the scene location where the loop occurs. We form a new segment $\mathcal{S}_L$ by concatenating images surrounding images $I \in \mathcal{S}_i$ and images $I' \in \mathcal{S}_j$. 
This segment $\mathcal{S}_L$ contains distant views of the same scene location and overlaps with segments $\mathcal{S}_i$ and $\mathcal{S}_j$. Processed by \muster{} model, this additional local reconstruction complements the sequential processing of adjacent segments and provides \slider{} with a more diverse, time-dispersed perspective, enabling a more robust scene reconstruction.

The 3D pointmap of segment $\mathcal{S}_L$ 
is then aligned with the pointmaps of the corresponding segments $\mathcal{S}_i$ and $\mathcal{S}_j$. We close the loop in the pose graph by chaining the alignments through the new segment $\mathcal{S}_L$ (see green node in Fig.~\ref{fig:graph}). 
We compute transforms to align segment $\mathcal{S}_i$ and segment $\mathcal{S}_L$, then $\mathcal{S}_L$ and $\mathcal{S}_j$. Similarly to the processing of adjacent segments, we complete the pose graph with two transforms, ${\bf T}_{i,L}^{x}$ and ${\bf T}_{i,L}^{p}$, which align $\mathcal{S}_i$ and $\mathcal{S}_L$, and two transforms, ${\bf T}_{L,j}^{x}$ and ${\bf T}_{L,j}^{p}$ which align $\mathcal{S}_L$ and $\mathcal{S}_j$.
They provide additional geometric constraints for the global optimization by bridging the two distant segments through an additional local 3D reconstruction.

Once the graph is completed, we can globally optimize all transforms in the pose graph~\cite{gigaslam,dpv-slam}. 
The pose graph is composed of adjacent and loop segments; built segment-wise, it is much smaller in the number of nodes and edges than complex frame-based factor graphs constructed by \master{}-SLAM and \vggt-SLAM. 
We minimize an objective function composed of two types of geometric constraints: sequential constraints from adjacent segments (Sec. \ref{sec:align}) and loop closure constraints from non-adjacent segments. 

This non-linear least-squares problem is efficiently solved using the Levenberg-Marquardt (LM) algorithm. By blending Gauss-Newton with gradient descent, the LM algorithm redistributes error over all nodes so that all constraints are satisfied as much as possible. It operates segment-wise and converges in few iterations, due to a small graph size.

\begin{table*}[h!]
    \vspace{-0.3em}
    \centering
    \caption{Root mean square error~(RMSE) of absolute pose error~(APE) on TUM RGB-D~\cite{Sturm12iros-TUM-RGB-D}.
    The gray rows indicate the results using the calibrated camera intrinsics; symbol '*' indicates baselines evaluated in the uncalibrated mode. Green is best and light green are second and third best.}
    \label{tab:tum_ate}
    \scriptsize
    \begin{tabular}{l|lcccccccccc} 
    \toprule
    \multirow{2}{*}{} & \multirow{2}{*}{Method} & \multicolumn{9}{c}{sequence} &  \multirow{2}{*}{Avg}  \\ \cmidrule(lr){3-11}
    & &\texttt{360} &\texttt{desk} &\texttt{desk2} &\texttt{floor} &\texttt{plant} &\texttt{room } &\texttt{rpy} &\texttt{teddy} &\texttt{xyz} & \\
    \midrule
    \cg
    \parbox[t]{1mm}{\multirow{8}{*}{\rotatebox[origin=c]{90}{Calibr.}}} 
    & \cg ORB-SLAM3~\cite{orbslam3} & $ \cg \times$ & \cg {0.017} & \cg 0.210 &$ \cg \times$ & \cg 0.034 &$ \cg  \times$ & \cg $\times$ &  \cg $\times$ & \cg \textbf{0.009} & \cg  N/A \\
    & \cg DeepV2D~\cite{Teed20iclr-DEEPV2D} & \cg 0.243 & \cg 0.166 & \cg 0.379 & \cg 1.653 & \cg 0.203 & \cg 0.246 & \cg 0.105 & \cg 0.316 & \cg 0.064 & \cg 0.375 \\
    & \cg DeepFactors~\cite{Czarnowski20ral-Deepfactors} & \cg 0.159 & \cg 0.170 & \cg 0.253 & \cg 0.169 & \cg 0.305 & \cg 0.364 & \cg 0.043 & \cg 0.601 & \cg 0.035 & \cg 0.233 \\
    & \cg DPV-SLAM~\cite{dpv-slam} & \cg 0.112 & \cg 0.018 & \cg 0.029 & \cg 0.057 & \cg 0.021 & \cg 0.330 & \cg 0.030 & \cg 0.084 & \cg {0.010} & \cg 0.076 \\
    & \cg DPV-SLAM++~\cite{dpv-slam} & \cg 0.132 & \cg 0.018 & \cg 0.029 & \cg 0.050 & \cg 0.022 & \cg 0.096 & \cg 0.032 & \cg 0.098 & \cg {0.010} & \cg 0.054 \\
    & \cg GO-SLAM~\cite{Zhang23iccv-GOSLAM} & \cg 0.089 & \cg \textbf{0.016} & \cg {0.028} & \cg {0.025} & \cg 0.026 & \cg {0.052} & \cg \textbf{0.019} & \cg 0.048 & \cg {0.010} & \cg {0.035} \\
    & \cg DROID-SLAM~\cite{Teed21nips-DROID-SLAM} & \cg 0.111 & \cg 0.018 & \cg 0.042 & \cg \textbf{0.021} & \cg \textbf{0.016} & \cg \textbf{0.049} & \cg {0.026} & \cg 0.048 & \cg 0.012 & \cg 0.038 \\
    & \cg \master-SLAM~\cite{mast3r-slam} & \cg \textbf{0.049} & \cg \textbf{0.016} & \cg \textbf{0.024} & \cg {0.025} & \cg {0.020} & \cg 0.061 & \cg 0.027 & \cg \textbf{0.041} & \cg \textbf{0.009} & \cg \textbf{0.030} \\
    \midrule
    \hline 
    \parbox[t]{1mm}{\multirow{4}{*}{\rotatebox[origin=c]{90}{Uncalibr.}}} 
    &DROID-SLAM*~\cite{Teed21nips-DROID-SLAM} 
              &        0.202 & \thirdc 0.032 &         0.091 & \thirdc 0.064 &         0.045 &         0.918 &         0.056 &\secondc 0.045 & \firstc 0.012 &          0.158 \\
    &\master-SLAM*~\cite{mast3r-slam} 
              &\thirdc 0.070 &        0.035 &         0.055 &\secondc 0.056 &\secondc 0.035 &         0.118 &         \thirdc 0.041 &         0.114 &         0.020 & \thirdc 0.060 \\
    &VGGT-SLAM, SIM(3) \cite{maggio2025vggtslamdensergbslam} 
              & 0.123 & 0.040 & 0.055 & 0.254 & \firstc 0.022 & \secondc 0.088 & \secondc0.041 & \firstc 0.032 & \thirdc 0.016 & 0.074 \\
    &VGGT-SLAM, SL(4) \cite{maggio2025vggtslamdensergbslam}
              & 0.071 & \firstc 0.025 & \firstc 0.040 &         0.141 & \firstc 0.023 &\thirdc 0.102 &\secondc 0.030 & \firstc 0.034 &\secondc 0.014 & \secondc 0.053 \\
    &VGGT-Long~\cite{vggt-long}&\secondc 0.063&         0.059 &\secondc 0.045 & 0.108 &         0.057 &         0.172 &         0.056 &         0.137 &         0.051 &             0.083 \\
    &\slider(ours)&\firstc 0.062 &\secondc 0.031 & \thirdc 0.047 & \firstc 0.054 &         0.054 & \firstc 0.065 & \firstc 0.028 & 0.115 &         0.018 & \firstc 0.052 \\
    \bottomrule
    \end{tabular}
    \vspace{-0.3em}
\end{table*}

\vspace{-0.3em}
\begin{table*}[h!]
    \centering
    \caption{Root mean square error~(RMSE) of absolute trajectory error (ATE) on 7-Scenes~\cite{Shotton13cvpr} (unit: m). 
    }
    \label{tab:7scenes_ate}
    \scriptsize
    \begin{tabular}{l|lcccccccc} 
    \toprule
    \multirow{2}{*}{} & \multirow{2}{*}{Method} & \multicolumn{7}{c}{Sequence} &  \multirow{2}{*}{Avg}  \\ \cmidrule(lr){3-9}
                      & &\texttt{chess} &\texttt{fire} &\texttt{heads} &\texttt{office} &\texttt{pumpkin} &\texttt{kitchen} &\texttt{stairs} & \\
    \midrule
    \cg
    \parbox[t]{1mm}{\multirow{3}{*}{\rotatebox[origin=c]{90}{Calibr.}}} 
        & \cg NICER-SLAM~\cite{Zhu243dv-NICERSLAM} & \cg \textbf{0.033} & \cg 0.069 & \cg 0.042 & \cg 0.108 & \cg 0.200 &  \cg \textbf{0.039} & \cg 0.108 & \cg 0.086 \\
        & \cg DROID-SLAM~\cite{Teed21nips-DROID-SLAM} & \cg {0.036} & \cg {0.027} & \cg {0.025} & \cg \textbf{0.066} & \cg 0.127 & \cg  {0.040} & \cg  {0.026} & \cg  {0.049} \\
        & \cg \master-SLAM~\cite{mast3r-slam} & \cg 0.053 & \cg \textbf{0.025} &  \cg \textbf{0.015} & \cg {0.097} & \cg  \textbf{0.088} & \cg 0.041 & \cg  \textbf{0.011} & \cg  \textbf{0.047} \\ \midrule
    \parbox[t]{1mm}{\multirow{4}{*}{\rotatebox[origin=c]{90}{Uncalibr.}}} 
        &DROID-SLAM*~\cite{Teed21nips-DROID-SLAM} 
                &\secondc 0.047&\secondc 0.038&       0.034 &        0.136 &       0.166 &        0.080 & \secondc 0.044&        0.078  \\
        &\master-SLAM*~\cite{mast3r-slam} 
                & 0.063     &        0.046 & 0.029 & \thirdc 0.103 &\firstc 0.114 &        0.074& \firstc 0.032&\firstc 0.066 \\                             
        &\vggt-SLAM, SIM(3)~\cite{maggio2025vggtslamdensergbslam} 
                & \secondc 0.037 & \firstc 0.026 & \firstc 0.018 &  0.104 & \secondc 0.133 & \thirdc0.061 & 0.093 & \secondc 0.067 \\
        &\vggt-SLAM, SL(4)~\cite{maggio2025vggtslamdensergbslam}  
                & \firstc 0.036 & \secondc 0.028 & \firstc 0.018 & \firstc 0.103 & \secondc 0.133 & \firstc0.058 & 0.093 & \secondc 0.067  \\
        &\vggt-Long~\cite{vggt-long}
                & 0.054 &        0.055 &       0.059 &         0.110 &        0.245& 0.067& \thirdc 0.055&        0.092  \\
        &\slider{}(ours)           
                & 0.054 & 0.041&\thirdc 0.028&\firstc 0.098 & 0.145&\firstc 0.053&         0.057&\secondc 0.067  \\
    \bottomrule
    \end{tabular}
\end{table*}

\section{Experiments}
\label{sec:exp}
We evaluate \slider{} on standard RGB SLAM benchmarks to assess the camera pose and angle estimation accuracies on the 7-Scenes~\cite{Shotton13cvpr} and TUM RGB-D~\cite{Sturm12iros-TUM-RGB-D} datasets; we report root mean square error~(RMSE) of the absolute pose error~(APE) using EVO metric~\cite{Grupp17evo}. We also evaluate the 3D reconstruction on a proprietary dataset of uncalibrated image sequences collected by navigational robots in a multi-room office environment.  
On 7-Scenes and TUM RGB-D datasets, we compare \slider{} with the state-of-the-art approaches in uncalibrated setting: DROID-SLAM~\cite{Teed21nips-DROID-SLAM}, \master-SLAM~\cite{mast3r-slam}, \vggt-SLAM~\cite{maggio2025vggtslamdensergbslam} and \vggt-Long~\cite{vggt-long}. We use reported numbers from \master-SLAM~\cite{mast3r-slam} and \vggt-SLAM~\cite{maggio2025vggtslamdensergbslam} for baselines, including the uncalibrated version of DROID-SLAM which estimates intrinsics with an automatic calibration pipeline~\cite{Veicht24eccv-Geocalib}. For the sake of comparison, we also include the state-of-the-art methods for the calibrated setting~\cite{orbslam3,Czarnowski20ral-Deepfactors,dpv-slam,Zhang23iccv-GOSLAM,Teed20iclr-DEEPV2D,Zhu243dv-NICERSLAM} provided with camera intrinsics. 

We use the MUSt3R$\_$512.pth pre-trained model with ViT-L encoder and ViT-B decoder available at {\tt https://github.com/naver/must3r}. For the fair comparison with the competitor \vggt-Long, we run \slider{} with the segment length $l$=60 and overlap $p$=30, and SIM(3) transform group. The cosine similarity threshold is $\sigma_{sim}=0.95$. In ablation studies (Sec.~\ref{sec:ablation}), we run \slider{} with other segmentation parameters and transform groups.

\subsection{Pose estimation evaluation}
\label{sec:pose}
As shown in Tabs.~\ref{tab:tum_ate} and \ref{tab:7scenes_ate}, \slider{} performs comparably to the top performing uncalibrated baselines on 7-Scenes and TUM RGB-D datasets. On 7-Scenes, \slider{} shows the same APE as the top performing baselines \master-SLAM and \vggt-SLAM, despite the lightweight loop closure and global optimization. On the TUM dataset, \slider{} performs the best overall with an average error of 0.052 m. At the same time, it outperforms by a large margin the competitor \vggt-Long on all sequences of the two datasets, under the same segmentation strategy. This demonstrates that we are able to extend \muster{} to multiple sequences with a rather simple pipeline of segmenting the input sequence and stitching local pointmaps.

In addition to the camera pose estimation with APE, we estimate the average angular error (AAE) and compare \slider{} to \vggt-Long{} on more complex \texttt{freiburg2} and \texttt{freiburg3} TUM scenes. As Tab.~\ref{tab:angular} shows, \slider{} generates more accurate estimation with the lower pose and angular errors than \vggt-Long{}, in the same segmentation setup (segment length $l$=60 and overlap $p$=30).  
%
\begin{table*}[ht!]
\centering
\caption{Average pose and angular errors of \vggtlong{} and \slider{}, on \texttt{freiburg2} and \texttt{freiburg3} TUM scenes. Green is the best.}
\label{tab:angular}
\scriptsize
\begin{tabular}{l|c|ccccccccccc}
\toprule
Method                    & Error       & \texttt{cabi}  &\texttt{large}  &\texttt{long\_off}&\texttt{ nostr}   & \texttt{nostr\_tex}&\texttt{sitting} &\texttt{sitting} &\texttt{sitting}&\texttt{walking}& \texttt{teddy} \\
                          &             & \texttt{net}   &\texttt{cabinet}& \texttt{house}   & \texttt{tex\_far}& \texttt{near\_wl} & \texttt{half}   & \texttt{rpy}    & \texttt{stat}    & \texttt{xyz}   &       \\ 
\midrule
VGGT-Long \cite{vggt-long}& APE & 0.095 & 0.127 & 0.167   &  0.156  & \firstc 0.154     & 0.211  & 0.061  & 0.014 & 0.169 & 0.170 \\ 
                          & AAE & \firstc 3.319 & 4.913 & 5.60    &   7.97  & \firstc 2.25      & 19.87  & 93.48  & \firstc 12.71 & 15.35 & 5.575 \\ \hline
\slider{}(ours)           & APE & \firstc 0.072 & \firstc 0.109 & \firstc 0.081   & \firstc  0.091 & 0.243     & \firstc 0.136  & \firstc 0.054  & \firstc 0.012 & \firstc 0.158 & \firstc 0.054 \\ 
\slider{}(ours)           & AAE & 7.41  & \firstc 2.92  & \firstc 2.48    & \firstc  22.77 & 11.38     & \firstc 10.35  & \firstc 84.64  & 16.97 & \firstc 13.04 & \firstc 5.335 \\ \hline
\end{tabular}
\vspace{-0.3em}
\end{table*}

Finally, we run \master-SLAM, \muster{}, \vggt-Long and \slider{} on the robot navigation collection. As Tab.\ref{tab:CC} shows, the first three methods work well except when robots navigate in a narrow corridor with featureless walls (see at the bottom of Fig. \ref{fig:3D-CC}) while \slider{} allows to recover the robot track thanks to segment overlaps and loop closures.

\vspace{-0.3em}
\begin{table}[ht!]
\scriptsize
\centering
\caption{Average pose and angular errors on robot navigation collection} %
\label{tab:CC}
\begin{tabular}{l|cc}  
\toprule
Method                            &        AAE  &          APE  \\
\midrule 
\master{}-SLAM \cite{mast3r-slam} &       29.47 &         1.580  \\           
\muster{} \cite{cabon25muster}    &       11.87 &         1.871  \\                      
\vggt-Long \cite{vggt-long}       &       33.31 &         2.239  \\           
\slider{} (ours)                  &\firstc 7.10 & \firstc 0.251 \\ \hline
\end{tabular}
\vspace{-0.3em}
\end{table}

\begin{figure*}[ht]
    \centering
    \includegraphics[width=0.245\textwidth]{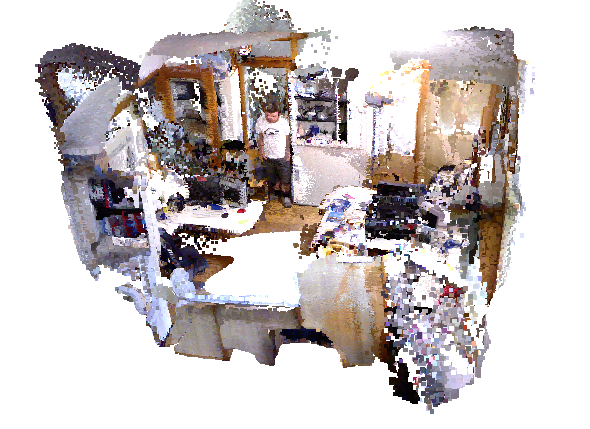}
    \includegraphics[width=0.245\textwidth]{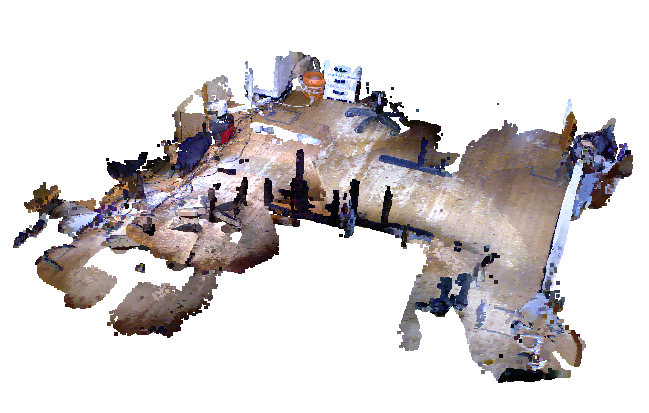}
    \includegraphics[width=0.245\textwidth]{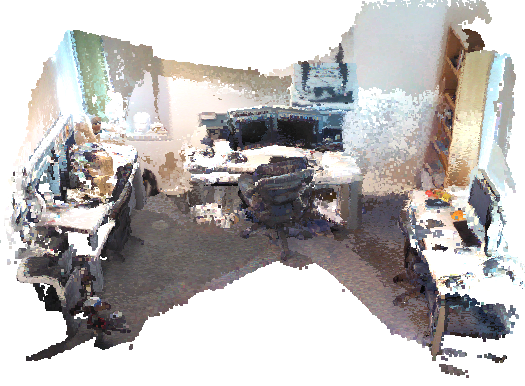}
    \includegraphics[width=0.245\textwidth]{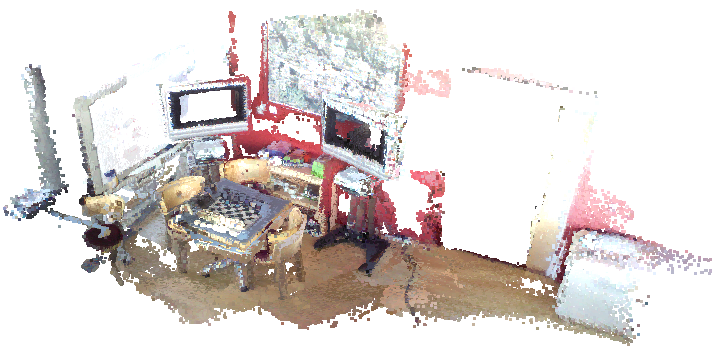}
    \vspace{-0.3em} 
    \includegraphics[width=0.245\textwidth]{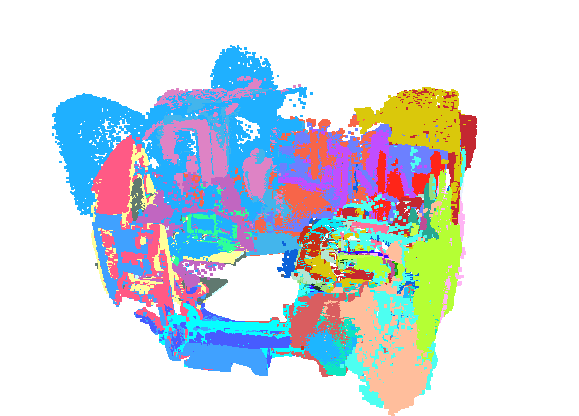}
    \includegraphics[width=0.245\textwidth]{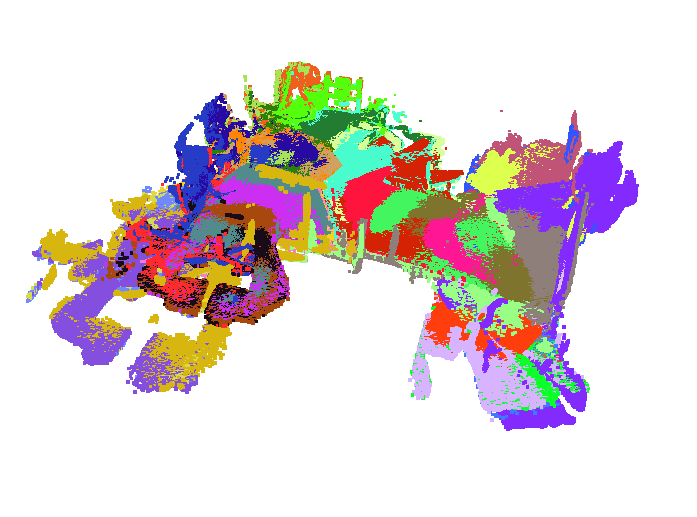}
    \includegraphics[width=0.245\textwidth]{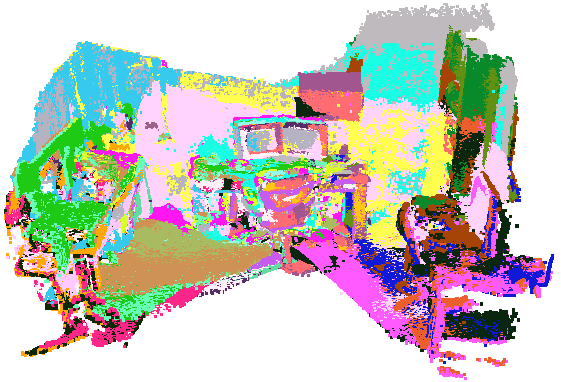}
    \includegraphics[width=0.245\textwidth]{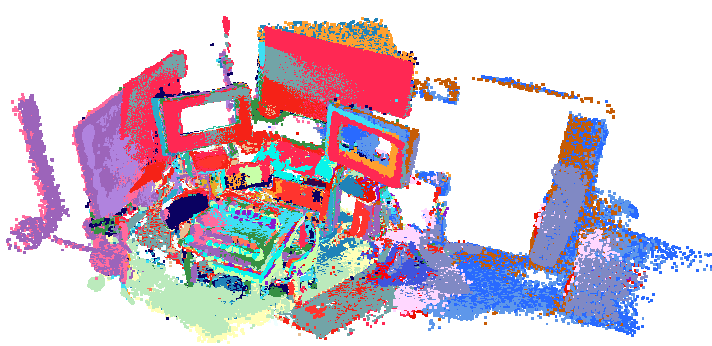}
    \caption{3D reconstruction with \slider{} for \texttt{room} and \texttt{floor} sequences of TUM dataset and \texttt{office} and \texttt{chess} of 7-Scenes dataset.
    Top: 3D reconstruction with original RGB colors; Bottom: segment pointmaps shown with different colors.}
    \label{fig:colored_segments}
\end{figure*}

\subsection{Ablations}
\label{sec:ablation}
We report ablation studies performed on \slider{} pipeline with conclusions summarized as follows: 
(a) the effect of different values of segment length $l$, where cutting the input sequence in longer sequences leads generally to higher accuracy on 3D reconstruction and pose estimation,
(b) comparison of three transform groups, where SIM(3) transform group represents a globally more reliable solution for aligning segments than Affine(3) and SL(4),
(c) the effect of loop closure and global optimization which leads to reducing the pose and angular errors, and estimating both alignments between two overlapping segments, using pointmaps ${\bf T}_{ij}^{x}$ and camera poses ${\bf T}_{ij}^{p}$, is the winner strategy.
\vspace{-0.3em}
\begin{table}[ht!]
\scriptsize
\centering
\caption{Ablation on segment length: AAE and APE for TUM and 7-Scenes. 
Green is the best, yellow refers to Tabs.~\ref{tab:tum_ate} and~\ref{tab:7scenes_ate}.}
\begin{tabular}{l|cc|cc}  
\toprule
Method & \multicolumn{2}{c}{TUM \texttt{teddy}} & \multicolumn{2}{c}{7-Scenes \texttt{pumpkin}} \\ \cline{2-5}
                  & AAE & APE  & AAE & APE \\
\midrule 
\slider{} $l$=20  & 9.59 & 0.185 & 3.88 & 0.156  \\           
\slider{} $l$=40  & 8.30 & 0.164 & 3.92 & 0.155  \\        
\slider{} $l$=60  & 7.66 & \thirdc 0.115 & 3.77 & \thirdc 0.145  \\  
\slider{} $l$=80  & 4.78 & 0.091 & 3.83 & 0.146  \\
\slider{} $l$=100 & 6.03 & 0.141 & \firstc 3.13 & 0.147 \\
\slider{} $l$=200 & 4.27 & 0.087 & 4.12 & 0.141 \\   
\muster{}~\cite{cabon25muster} & \firstc 1.42 & \firstc 0.061 & 3.36 &  \firstc 0.137 \\ \hline
\end{tabular}
\label{tab:segmentsize}
\end{table}

\myparagraph{Segment length}
Tab.~\ref{tab:segmentsize} ablates the segment length for two sequences from 7-Scenes and TUM, with the segment size $l$ varying from 20 to 200 and the overlap $p$ fixed to $l/2$. In both sequences, partitioning the input sequence in longer segments and stitching fewer local pointmaps help to reduce both average angular and pose errors. They provide a substantial gain over the default case $l=60$ and bring \slider{} closer to the results achieved of \muster{} 
that processes the full sequence in one pass.

\myparagraph{Transform groups}
All \slider{} results in Tabs.~\ref{tab:tum_ate} and ~\ref{tab:7scenes_ate} are reported using SIM(3) as the transformation group. We additionally evaluated our pipeline with two alternative groups: Affine(3) and SL(4). Across all experiments, SIM(3) consistently demonstrated its strength by delivering fast and reliable estimates of both pointmaps and camera poses. Substituting SIM(3) with the more expressive Affine(3) group led to only a modest improvement, reducing APE by about 0.5\%. In contrast, the SL(4) group showed mixed performance: while it achieved up to a 4\% APE reduction in three sequences, it failed to produce stable alignments in two others, requiring additional per-scene parameter tuning. These results confirm the sensibility of SL(4) to outliers and its tendency toward unstable homographic solutions in the presence of planar structures, an issue also noted in ~\cite{maggio2025vggtslamdensergbslam}.

Another key consideration is computation time. For short segments, transform estimation times are comparable across all groups. However, once the segment length $l$ exceeds 100, the cost for Affine(3) and especially SL(4) increases rapidly. As a result, \slider{} is forced to heavily sparsify the pointmaps on longer segments before applying SL(4) transforms, effectively negating the potential benefits of a more expressive transformation group. Overall, within our pipeline, SIM(3) proves to be the most reliable and balanced choice among the three.

\myparagraph{Metric vs non-metric predictions}
As mentioned before, \slider{} inherits the accurate metric output from \muster{}. 
Error of scale estimation of \slider{} is less that 2\% on both TUM and 7-Scenes. 
This makes predictions directly deployable for the robot navigation, while \vggt-Long predictions require other sources of information for rescaling before acting in robot navigation. In Tabs.~\ref{tab:tum_ate} and ~\ref{tab:7scenes_ate} results for both \vggt-Long and \slider{} are reported after rescaling.

\myparagraph{Loop closure and global optimization}
Despite the lightweight segment-wise approach, the loop closure and optimization 
reduces APE and AAE. The average gain on 7-Scenes is 14\% and 3\%, respectively.
Fig.~\ref{fig:traj-optim} shows trajectory of TUM and 7-Scenes sequences, before and after optimization. Moreover, estimating two alignments, from pointmaps and camera poses, is beneficial; dropping the former reduces the gain by 4\% and the later by 3\%.

\vspace{-0.3em} 
\subsection{Qualitative results}
\label{sec:qualitative}
In this section, we present qualitative results to illustrate the quality of scene reconstruction by \slider{}.  

Fig.~\ref{fig:colored_segments} (top) shows examples of 3D scene reconstruction with original RGB colors of \texttt{floor} and \texttt{desk} scenes from TUM RGB-D dataset, and \texttt{office} and \texttt{chess} from 7-Scenes. Additionally, Fig.~\ref{fig:colored_segments} (bottom) shows the same scenes with all segments painted with different colors; this help to visualize the alignments between segments and contribution of each segment in the final scene reconstruction.

\begin{figure*}[h!]
    \centering
    \includegraphics[width=0.136\textwidth]{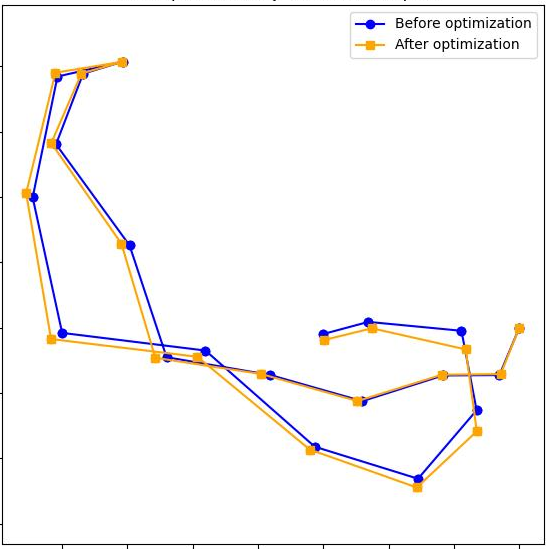}
    \includegraphics[width=0.136\textwidth]{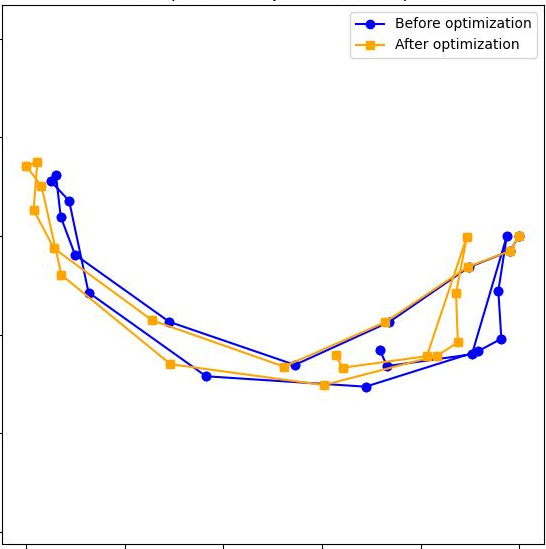}
    \includegraphics[width=0.136\textwidth]{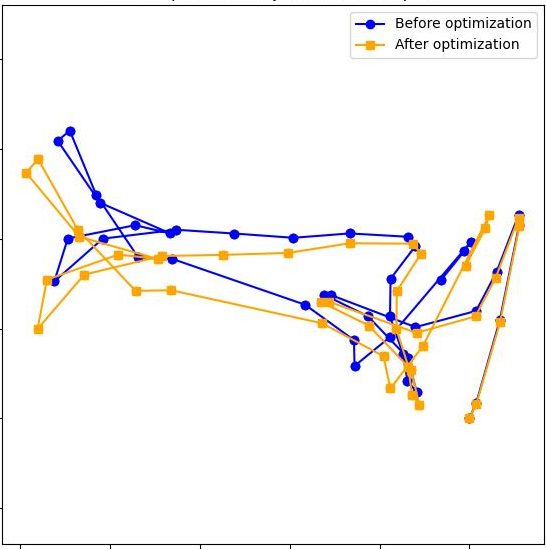}
    \includegraphics[width=0.136\textwidth]{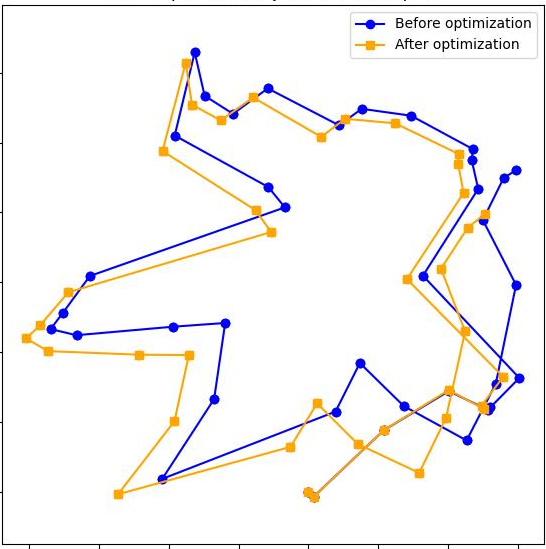}
    \includegraphics[width=0.136\textwidth]{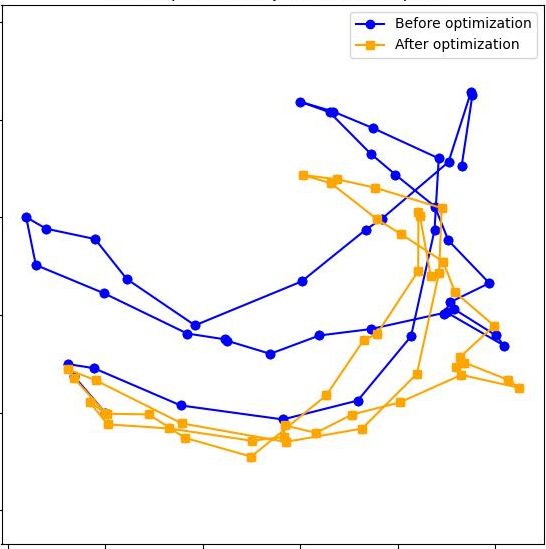}
    \includegraphics[width=0.136\textwidth]{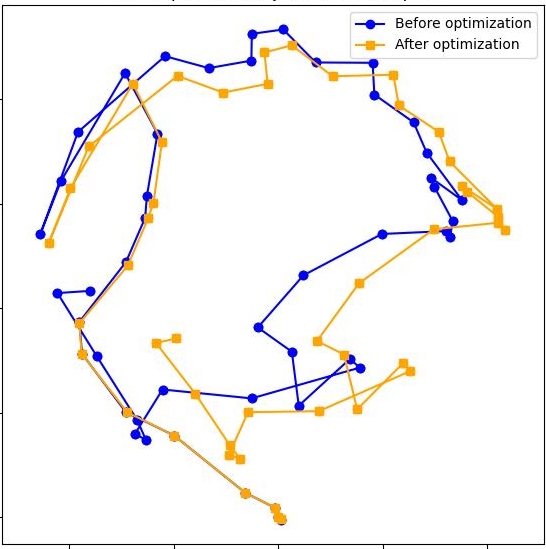}
    \includegraphics[width=0.136\textwidth]{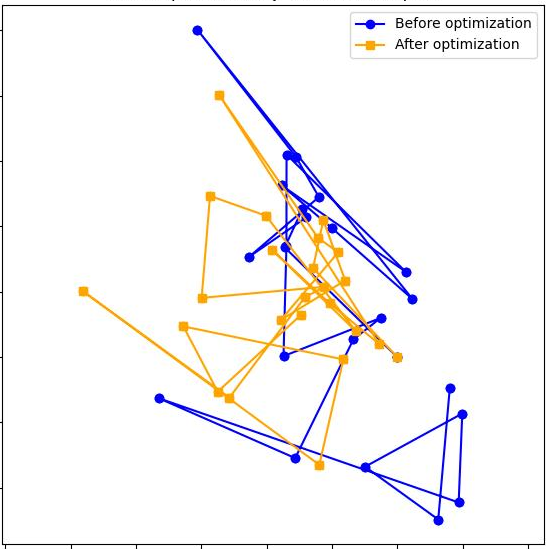}
    \vspace{-0.1em} 
    \includegraphics[width=0.136\textwidth]{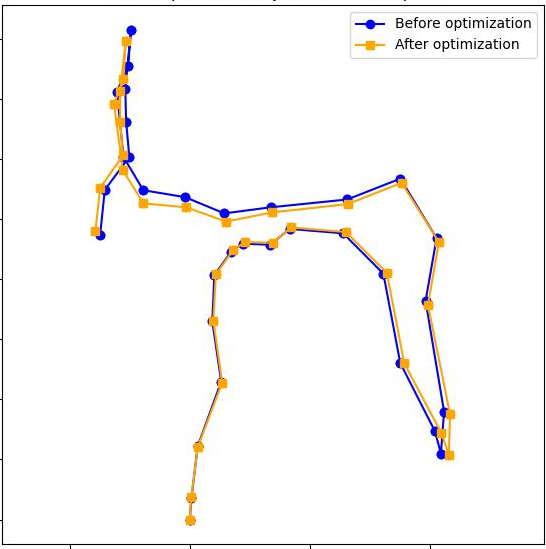}
    \includegraphics[width=0.136\textwidth]{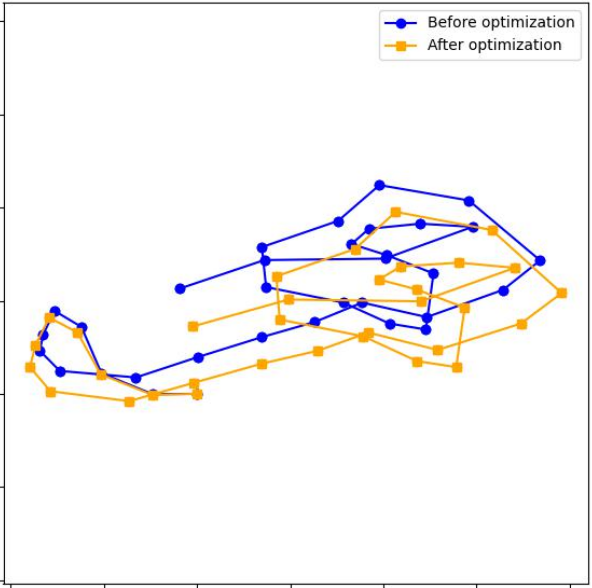}
    \includegraphics[width=0.136\textwidth]{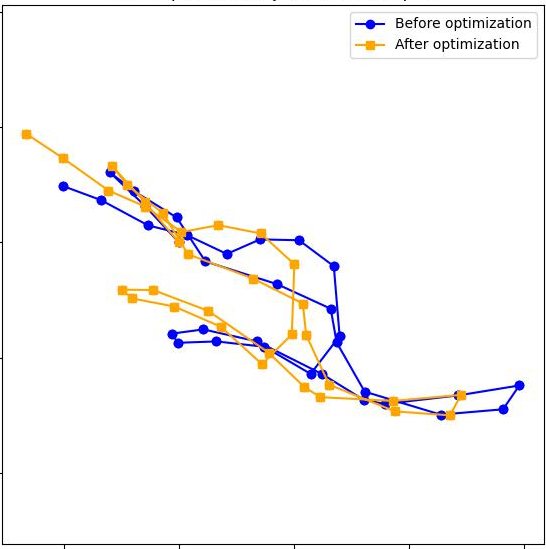}
    \includegraphics[width=0.136\textwidth]{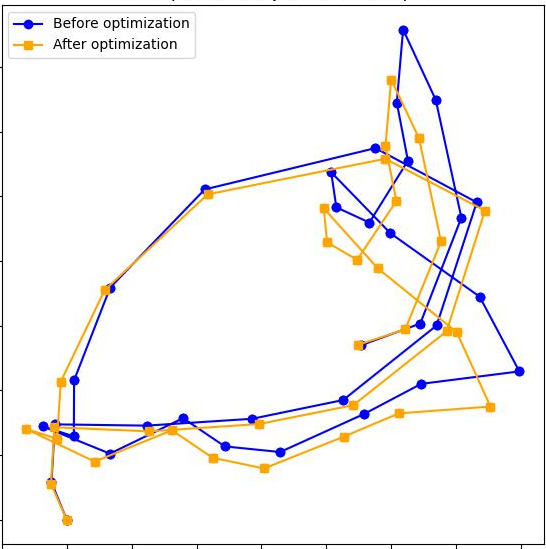}
    \includegraphics[width=0.136\textwidth]{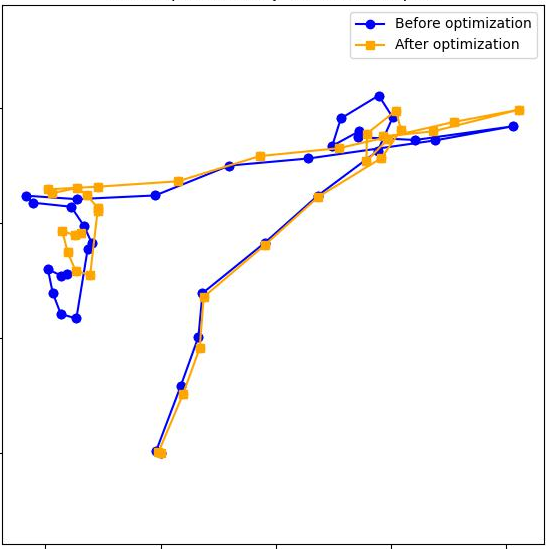}
    \includegraphics[width=0.136\textwidth]{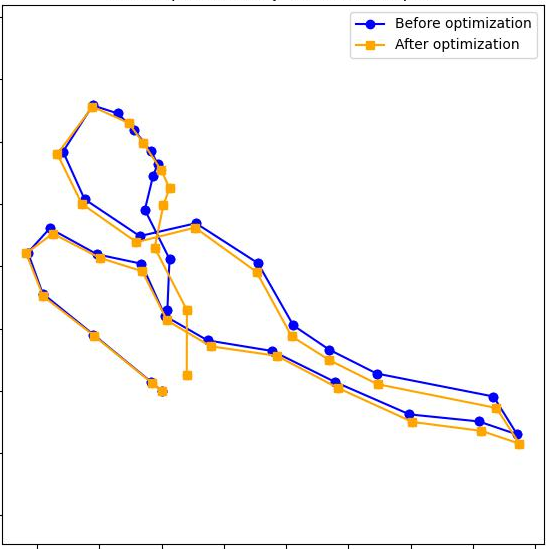}
    \includegraphics[width=0.136\textwidth]{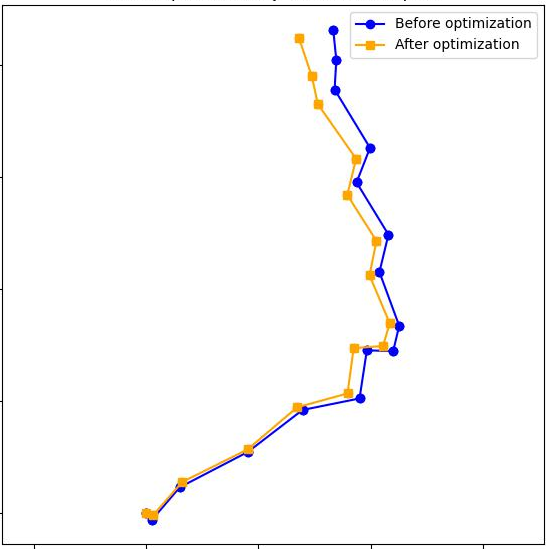}
    \caption{Segment pose correction with lightweight loop closure, for TUM (top) and 7-Scenes (down) datasets.}
    \label{fig:traj-optim}
\end{figure*}

\begin{figure*}[h!]
    \centering
    \includegraphics[width=0.28\textwidth]{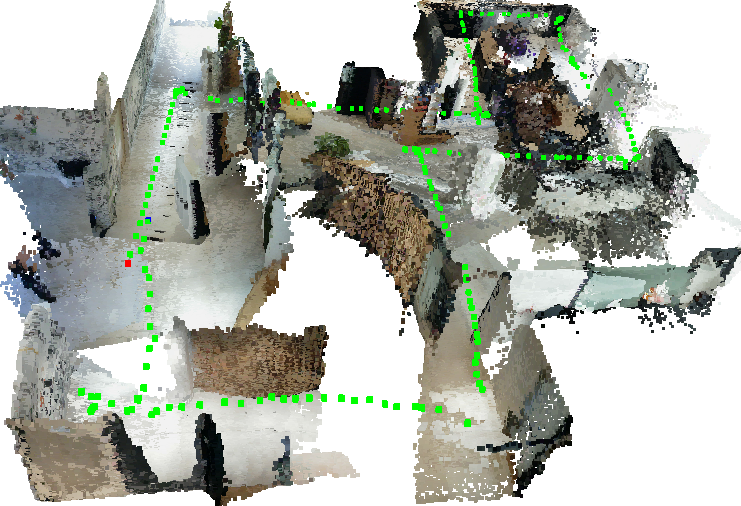} 
    \quad \quad \quad
    \includegraphics[width=0.28\textwidth]{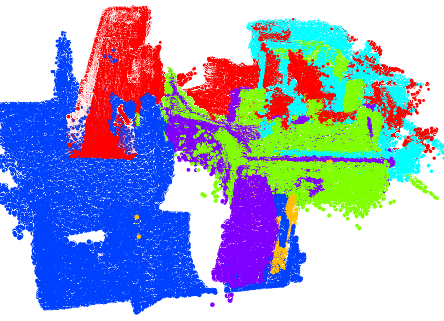}
    \quad \quad \quad
    \includegraphics[width=0.28\textwidth]{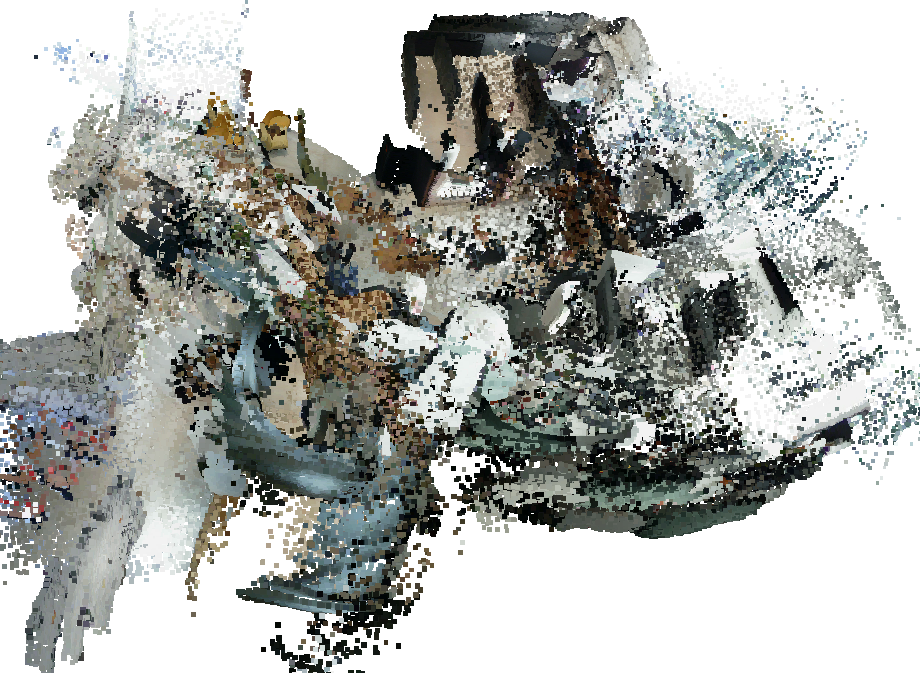}
    \caption{3D reconstruction from a robot navigation sequence with \slider{} in RGB (left) and segment-colored (middle) versions and with \vggt-Long (right).}
    \label{fig:3D-CC}
    \vspace{-0.3em}
\end{figure*}

Fig.~\ref{fig:traj-optim} demonstrates the effect of using lightweight loop closure optimization on correcting segment poses in TUM and 7-Scenes datasets.

Fig.~\ref{fig:3D-CC} illustrates scene reconstruction from a robot navigation sequence. Figs.~\ref{fig:3D-CC}(left) and (middle) show the pointmap generated by \slider{} with the estimated robot trajectory and its segment-colored version.
Instead, the scene reconstruction with \vggt-Long in Fig.~\ref{fig:3D-CC}(right) is much less accurate. For better visualization, both 3D representations have been modified by removing the ceiling of the scene.

\subsection{Limitations} 
\label{sec:limitations}
A clear limitation of the stitching approach is its strong dependence on the quality of local reconstructions produced by the foundation model. While \slider{} shows some ability to mitigate these errors, by leveraging segment overlaps as an additional source of information to filter out inaccuracies from the \muster{} model, this 
effect remains limited in scope.

Another limitation, shared with most SLAM methods that rely on loop closure to correct drift, is the high sensitivity to hyperparameters such as the similarity threshold $\sigma_{sim}$. More robust alternatives, including self-tuning mechanisms, adaptive acceptance thresholds, or probabilistic formulations, could be integrated to improve outlier detection and enhance the reliability of loop closure.

\vspace{-0.3em} 
\section{Conclusions} 
\label{sec:conclusion}
In this paper, we introduce \slider{}, a simple yet effective pipeline that extends the capabilities of the \muster{} foundation model for monocular 3D reconstruction from uncalibrated images. The method tackles scalability by segmenting input sequences, aligning the resulting segments, and applying lightweight loop closure optimization. Several design improvements are incorporated to enhance accuracy and consistency, yielding more robust alignments and reconstructions.
Experiments on the 7-Scenes, TUM and robot navigation datasets demonstrate that \slider{} achieves consistent 3D reconstructions from long RGB sequences. These results underline the potential of leveraging the \muster{} model for scalable monocular 3D scene reconstruction in real-world scenarios, with the key advantage of producing predictions directly in metric space.

\bibliographystyle{plain}
\bibliography{slider}

@string{ijcv    = {{International	Journal of Computer Vision}}}

@string{is      = {{Information Science (Elsevier)}}}

@string{pami    = {{IEEE Transactions on Pattern Analysis and Machine Intelligence}}}

@string{cvpr    = {{CVPR}}}

@string{eccv    = {{ECCV}}}

@string{iclr	= {{ICLR}}}

@string{icra	= {{ICRA}}}

@string{iros	= {{IROS}}}

@string{nips	= {{NeurIPS}}}

@InProceedings{master,
	Title		= {{Grounding Image Matching in 3D with MASt3R}},
	Author		= {Leroy, Vincent and Cabon, Yohann and Revaud, J\'{e}r\^{o}me},
	Booktitle	= {European Conference on Computer Vision},
	Year		= {2024}}

@misc{WangX24Spanner3DReconstructionWithSpatialMemory,
	Title		= {{3D Reconstruction with Spatial Memory}},
	Author		= {Hengyi Wang and Lourdes Agapito},
	Howpublished= {arXiv:2408.16061},
	Year		= {2024}}

@InProceedings{WangCVPR24VisualGeometryGroundedDeepSfM,
	Author		= {Wang, Jianyuan and Karaev, Nikita and Rupprecht, Christian and Novotny, David},
	Title		= {{Visual Geometry Grounded Deep Structure From Motion}},
	Booktitle	= {Proceedings of the Computer Vision and Pattern Recognition Conference},
	Year		= {2024}}

@misc{zhang24monster,
	Title		= {{MonST3R: A Simple Approach for Estimating Geometry in the Presence of Motion}},
	Author		= {Zhang, Junyi and Herrmann, Charles and Hur, Junhwa and Jampani, Varun and Darrell, Trevor and Cole, Forrester and Sun, Deqing and Yang, Ming-Hsuan},
	Howpublished= {arXiv:2410.03825},
	Year		= {2024}}

@inproceedings{eth3d_slam,
  author       = {Thomas Sch{\"{o}}ps and
                  Torsten Sattler and
                  Marc Pollefeys},
  title        = {{BAD SLAM: Bundle Adjusted Direct {RGB-D} SLAM}},
  booktitle    = {Proceedings of the Computer Vision and Pattern Recognition Conference},
  year         = {2019}
}

@inproceedings{smith24flowmap,
      title={{FlowMap: High-Quality Camera Poses, Intrinsics, and Depth via Gradient Descent}},
      author={Cameron Smith and David Charatan and Ayush Tewari and Vincent Sitzmann},
      year={2025},
      booktitle={3DV},
}

@article{deeptam,
  author       = {Huizhong Zhou and
                  Benjamin Ummenhofer and
                  Thomas Brox},
  title        = {{DeepTAM: Deep Tracking and Mapping with Convolutional Neural Networks}},
  journal      = ijcv,
  volume       = {128},
  number       = {3},
  pages        = {756--769},
  year         = {2020},}

@Book{hartleymultiviewgeometry,
	Title		= {{Multiple View Geometry in Computer Vision}},
	Author		= {Hartley, Richard and Zisserman, Andrew},
	Year		= {2004},
	Publisher	= {{Cambridge University Press}}}

@inproceedings{jiang13,
  author       = {Nianjuan Jiang and
                  Zhaopeng Cui and
                  Ping Tan},
  title        = {{A Global Linear Method for Camera Pose Registration}},
  booktitle    = {Proceedings of the International Conference on Computer Vision},
  year         = {2013}}

@Article{disco2013pami,
	Title		= {{{SfM with MRFs}: Discrete-Continuous Optimization for Large-Scale Structure from Motion}},
	Author		= {Crandall, David and Owens, Andrew and Snavely, Noah and Huttenlocher, Dan},
	Journal		= pami,
	Volume		= {35},
	Number		= {12},
	Pages		= {2841--2853},
	Year		= {2013},
	Publisher	= {IEEE}}

@inproceedings{colmapsfm,
    author={Sch\"{o}nberger, Johannes Lutz and Frahm, Jan-Michael},
    title={Structure-from-Motion Revisited},
    booktitle={Proceedings of the Computer Vision and Pattern Recognition Conference},
    year={2016}}

@inproceedings{shapefromsingleimg23,
      title={{Shape, Pose, and Appearance from a Single Image via Bootstrapped Radiance Field Inversion}}, 
      author={Dario Pavllo and David Joseph Tan and Marie-Julie Rakotosaona and Federico Tombari},
      booktitle={Proceedings of the Computer Vision and Pattern Recognition Conference},
      year={2023}}

@inproceedings{zero-1-to-3,
	title = {{Zero-1-to-3: {Zero}-shot {One} Image to {3D} Object}},
	author = {Liu, Ruoshi and Wu, Rundi and Van Hoorick, Basile and Tokmakov, Pavel and Zakharov, Sergey and Vondrick, Carl},
	booktitle = {Proceedings of the Computer Vision and Pattern Recognition Conference},
	year = {2023}}

@inproceedings{deepv2d,
  author       = {Zachary Teed and
                  Jia Deng},
  title        = {{DeepV2D: Video to Depth with Differentiable Structure from Motion}},
  booktitle    = iclr,
  year         = {2020}}

@article{vggt-long,
      title={VGGT-Long: Chunk it, Loop it, Align it -- Pushing VGGT's Limits on Kilometer-scale Long RGB Sequences}, 
      author={Kai Deng and Zexin Ti and Jiawei Xu and Jian Yang and Jin Xie},
      year={2025},
      journal={arXiv preprint arXiv:2507.16443},
}

@article{maggio2025vggtslamdensergbslam,
      title={VGGT-SLAM: Dense RGB SLAM Optimized on the SL(4) Manifold}, 
      author={Dominic Maggio and Hyungtae Lim and Luca Carlone},
      year={2025},
      journal={arXiv preprint arXiv:2505.12549},
}

@article{yu2025relativeposeestimationaffine,
      title={Relative Pose Estimation through Affine Corrections of Monocular Depth Priors}, 
      author={Yifan Yu and Shaohui Liu and Rémi Pautrat and Marc Pollefeys and Viktor Larsson},
      year={2025},
      journal={arXiv preprint arXiv:2501.05446},
      archivePrefix={arXiv},
      primaryClass={cs.CV},
      url={https://arxiv.org/abs/2501.05446}, 
}

@article{Cramariuc22ral-Maplab,
  author  = {Andrei Cramariuc and Lukas Bernreiter and Florian Tschopp and Marius Fehr and Victor Reijgwart and Juan Nieto and Roland Siegwart and Cesar Cadena},
  title   = {{maplab 2.0--A Modular and Multi-Modal Mapping Framework}},
  journal = {{IEEE} Robotics and Automation Letters},
  volume  = {8},
  number  = {2},
  pages   = {520--527},
  year    = {2022}
}

@article{Czarnowski20ral-Deepfactors,
  author  = {J. Czarnowski and T. Laidlow and R. Clark and A. Davison},
  title   = {{DeepFactors}: Real-Time Probabilistic Dense Monocular {SLAM}},
  journal = {{IEEE} Robotics and Automation Letters},
  volume  = {5},
  number  = {2},
  pages   = {721--728},
  year    = {2020}
}

@inproceedings{Furukawa10cvpr,
  author    = {Y. Furukawa and B. Curless and S. M. Seitz and R. Szeliski},
  title     = {Towards Internet-Scale Multi-View Stereo},
  booktitle = {IEEE Conf. on Computer Vision and Pattern Recognition (CVPR)},
  pages     = {1434--1441},
  year      = {2010}
}

@misc{Grupp17evo,
  author       = {Michael Grupp},
  title        = {{evo}: Python Package for the Evaluation of Odometry and {SLAM}},
  howpublished = {url{https://github.com/MichaelGrupp/evo}},
  year         = {2017}
}

@article{Jang25arxiv-Pow3R,
  author  = {Wonbong Jang and Philippe Weinzaepfel and Vincent Leroy and Lourdes Agapito and Jerome Revaud},
  title   = {{Pow3R: Empowering Unconstrained 3D Reconstruction with Camera and Scene Priors}},
  journal = {arXiv preprint arXiv:2503.17316},
  year    = {2025}
}

@article{Koenderink91,
  author  = {J. J. Koenderink and A. J. van Doorn},
  title   = {Affine Structure from Motion},
  journal = {Journal of the Optical Society of America A},
  volume  = {8},
  number  = {2},
  pages   = {377--385},
  year    = {1991}
}

@inproceedings{Pan24eccv-GLOMAP,
  author    = {Linfei Pan and Daniel Barath and Marc Pollefeys and Johannes Lutz Sch{\"o}nberger},
  title     = {{Global Structure-from-Motion Revisited}},
  booktitle = {European Conf. on Computer Vision (ECCV)},
  year      = {2024}
}

@inproceedings{Rosinol20icra-Kimera,
  author    = {A. Rosinol and M. Abate and Y. Chang and L. Carlone},
  title     = {Kimera: An Open-Source Library for Real-Time Metric-Semantic Localization and Mapping},
  booktitle = {{IEEE} Intl. Conf. on Robotics and Automation (ICRA)},
  pages     = {1689--1696},
  year      = {2020},
  note      = {arXiv preprint: 1910.02490}
}

@inproceedings{Schonberger16cvpr-SfMRevisited,
  author    = {Johannes L. Schonberger and Jan-Michael Frahm},
  title     = {Structure-from-Motion Revisited},
  booktitle = {IEEE Conf. on Computer Vision and Pattern Recognition (CVPR)},
  pages     = {4104--4113},
  year      = {2016}
}

@inproceedings{Shotton13cvpr,
  author    = {J. Shotton and B. Glocker and C. Zach and S. Izadi and A. Criminisi and A. Fitzgibbon},
  title     = {Scene Coordinate Regression Forests for Camera Relocalization in {RGB-D} Images},
  booktitle = {IEEE Conf. on Computer Vision and Pattern Recognition (CVPR)},
  pages     = {2930--2937},
  year      = {2013}
}

@inproceedings{Sturm12iros-TUM-RGB-D,
  author    = {J{\"u}rgen Sturm and Nikolas Engelhard and Felix Endres and Wolfram Burgard and Daniel Cremers},
  title     = {A Benchmark for the Evaluation of {RGB-D} {SLAM} Systems},
  booktitle = {{IEEE/RSJ} Intl. Conf. on Intelligent Robots and Systems (IROS)},
  pages     = {573--580},
  publisher = {{IEEE}},
  year      = {2012}
}

@inproceedings{Teed20iclr-DEEPV2D,
  author    = {Zachary Teed and Jia Deng},
  title     = {{DEEPV2D: Video to Depth with Differentiable Structure from Motion}},
  booktitle = {Intl. Conf. on Learning Representations (ICLR)},
  year      = {2018}
}

@inproceedings{Teed21nips-DROID-SLAM,
  author    = {Zachary Teed and Jia Deng},
  title     = {{DROID-SLAM}: Deep Visual {SLAM} for Monocular, Stereo, and {RGB-D} Cameras},
  booktitle = {Advances in Neural Information Processing Systems (NIPS)},
  year      = {2021}
}

@inproceedings{Veicht24eccv-Geocalib,
  author    = {Alexander Veicht and Paul-Edouard Sarlin and Philipp Lindenberger and Marc Pollefeys},
  title     = {{GeoCalib: Learning Single-Image Calibration with Geometric Optimization}},
  booktitle = {European Conf. on Computer Vision (ECCV)},
  pages     = {1--20},
  publisher = {Springer},
  year      = {2024}
}

@inproceedings{Zhang23iccv-GOSLAM,
  author    = {Youmin Zhang and Fabio Tosi and Stefano Mattoccia and Matteo Poggi},
  title     = {{GO-SLAM: Global Optimization for Consistent 3D Instant Reconstruction}},
  booktitle = {{Proceedings of the International Conference on Computer Vision}},
  pages     = {3727--3737},
  year      = {2023}
}

@inproceedings{Zhu243dv-NICERSLAM,
  author    = {Zihan Zhu and Songyou Peng and Viktor Larsson and Zhaopeng Cui and Martin R Oswald and Andreas Geiger and Marc Pollefeys},
  title     = {{NICER-SLAM: Neural Implicit Scene Encoding for {RGB} {SLAM}}},
  booktitle = {{IEEE} International Conference on 3D Vision (3DV)},
  pages     = {42--52},
  year      = {2024}
}

@inproceedings{cabon25muster,
  author       = {Yohann Cabon and Lucas Stoffl and Leonid Antsfeld and
                  Gabriela Csurka and Boris Chidlovskii and
                  J{\'{e}}r{\^{o}}me Revaud and Vincent Leroy},
  title        = {{MUSt3R: Multi-view Network for Stereo 3D Reconstruction}},
  booktitle    = {{Proceedings of the Computer Vision and Pattern Recognition Conference}},
  pages        = {1050--1060},
  publisher    = {Computer Vision Foundation / {IEEE}},
  year         = {2025},
}

@inproceedings{faster,
  author       = {Jianing Yang and
                  Alexander Sax and
                  Kevin J. Liang and
                  Mikael Henaff and
                  Hao Tang and
                  Ang Cao and
                  Joyce Chai and
                  Franziska Meier and
                  Matt Feiszli},
  title        = {Fast3R: Towards 3D Reconstruction of 1000+ Images in One Forward Pass},
  booktitle    = {{Proceedings of the Computer Vision and Pattern Recognition Conference}},
  pages        = {21924--21935},
  year         = {2025},
}

@inproceedings{mast3r-slam,
  title={MASt3R-SLAM: Real-time dense SLAM with 3D reconstruction priors},
  author={Murai, Riku and Dexheimer, Eric and Davison, Andrew J},
  booktitle={Proceedings of the Computer Vision and Pattern Recognition Conference},
  pages={16695--16705},
  year={2025}
}

@inproceedings{mast3r-sfm,
  title={MASt3R-SfM: a Fully-Integrated Solution for Unconstrained Structure-from-Motion},
  author={Duisterhof, Bardienus and Zust, Lojze and Weinzaepfel, Philippe and Leroy, Vincent and Cabon, Yohann and Revaud, Jerome},
  booktitle={{IEEE} International Conference on 3D Vision (3DV)},
  year={2025}
}

@article{cut3r,
  title={Continuous 3D Perception Model with Persistent State},
  author={Wang, Qianqian and Zhang, Yifei and Holynski, Aleksander and Efros, Alexei A and Kanazawa, Angjoo},
  journal={arXiv preprint arXiv:2501.12387},
  year={2025}
}

@inproceedings{vggt,
  title={Vggt: Visual geometry grounded transformer},
  author={Wang, Jianyuan and Chen, Minghao and Karaev, Nikita and Vedaldi, Andrea and Rupprecht, Christian and Novotny, David},
  booktitle={Proceedings of the Computer Vision and Pattern Recognition Conference},
  pages={5294--5306},
  year={2025}
}

@article{gigaslam,
  title={GigaSLAM: Large-Scale Monocular SLAM with Hierarchical Gaussian Splats},
  author={Deng, Kai and Zhang, Yigong and Yang, Jian and Xie, Jin},
  journal={arXiv preprint arXiv:2503.08071},
  year={2025}
}

@article{dfvo,
  title={DF-VO: What should be learnt for visual odometry?},
  author={Zhan, Huangying and Weerasekera, Chamara Saroj and Bian, Jia-Wang and Garg, Ravi and Reid, Ian},
  journal={arXiv preprint arXiv:2103.00933},
  year={2021}
}

@inproceedings{dvso,
  title={Deep virtual stereo odometry: Leveraging deep depth prediction for monocular direct sparse odometry},
  author={Yang, Nan and Wang, Rui and Stuckler, Jorg and Cremers, Daniel},
  booktitle={Proceedings of the European conference on computer vision (ECCV)},
  pages={817--833},
  year={2018}
}

@article{vaswani2017attention,
  title={Attention is all you need},
  author={Vaswani, Ashish and Shazeer, Noam and Parmar, Niki and Uszkoreit, Jakob and Jones, Llion and Gomez, Aidan N and Kaiser, {\L}ukasz and Polosukhin, Illia},
  journal={Advances in neural information processing systems},
  volume={30},
  year={2017}
}

@inproceedings{dao2023flashattention2,
  title={Flash{A}ttention-2: Faster Attention with Better Parallelism and Work Partitioning},
  author={Dao, Tri},
  booktitle={International Conference on Learning Representations (ICLR)},
  year={2024}
}

@article{cvivsic2022soft2,
  title={Soft2: Stereo visual odometry for road vehicles based on a point-to-epipolar-line metric},
  author={Cvi{\v{s}}i{\'c}, Igor and Markovi{\'c}, Ivan and Petrovi{\'c}, Ivan},
  journal={IEEE Transactions on Robotics},
  volume={39},
  number={1},
  pages={273--288},
  year={2022},
  publisher={IEEE}
}

@inproceedings{zhang2015visual,
  title={Visual-lidar odometry and mapping: Low-drift, robust, and fast},
  author={Zhang, Ji and Singh, Sanjiv},
  booktitle={2015 IEEE international conference on robotics and automation (ICRA)},
  pages={2174--2181},
  year={2015},
  organization={IEEE}
}

@article{zhang2014loam,
  title={LOAM: Lidar odometry and mapping in real-time.},
  author={Zhang, Ji and Singh, Sanjiv and others},
  booktitle={Robotics: Science and systems},
  volume={2},
  number={9},
  pages={1--9},
  year={2014},
  organization={Berkeley, CA}
}

@article{orbslam,
	doi = {10.1109/tro.2015.2463671},
	url = {    https://doi.org/10.1109/TRO.2015.2463671
    },
	year = 2015,
	month = {oct},
	publisher = {Institute of Electrical and Electronics Engineers ({IEEE})},
	volume = {31},
	number = {5},
	pages = {1147--1163},
	author = {Raul Mur-Artal and J. M. M. Montiel and Juan D. Tardos},
	title = {{ORB}-{SLAM}: A Versatile and Accurate Monocular {SLAM} System},
	journal = {{IEEE} Transactions on Robotics}
}

@article{orbslam2,
	doi = {10.1109/tro.2017.2705103},
	url = {https://doi.org/10.1109%2Ftro.2017.2705103},
	year = 2017,
	month = {oct},
	publisher = {Institute of Electrical and Electronics Engineers ({IEEE})},
	volume = {33},
	number = {5},
	pages = {1255--1262},
	author = {Raul Mur-Artal and Juan D. Tardos},
	title = {{ORB}-{SLAM}2: An Open-Source {SLAM} System for Monocular, Stereo, and {RGB}-D Cameras},
	journal = {{IEEE} Transactions on Robotics}
    }

@article{orbslam3,
	doi = {10.1109/tro.2021.3075644},
	url = {https://doi.org/10.1109%2Ftro.2021.3075644},
	year = 2021,
	month = {dec},
	publisher = {Institute of Electrical and Electronics Engineers ({IEEE})},
	volume = {37},
	number = {6},
	pages = {1874--1890},
	author = {Carlos Campos and Richard Elvira and Juan J. Gomez Rodriguez and Jose M. M. Montiel and Juan D. Tardos},
	title = {{ORB}-{SLAM}3: An Accurate Open-Source Library for Visual, Visual{\textendash}Inertial, and Multimap {SLAM}
    },
    journal = {{IEEE} Transactions on Robotics}
    }

@article{gradslam,
  title={gradSLAM: Automagically differentiable SLAM},
  author={Jatavallabhula, Krishna Murthy and Saryazdi, Soroush and Iyer, Ganesh and Paull, Liam},
  journal={arXiv preprint arXiv:1910.10672},
  year={2019}
}

@inproceedings{dpv-slam,
  title={Deep patch visual slam},
  author={Lipson, Lahav and Teed, Zachary and Deng, Jia},
  booktitle={European Conference on Computer Vision},
  pages={424--440},
  year={2024}
}

@inproceedings{wang2024dust3r,
  title={Dust3r: Geometric 3d vision made easy},
  author={Wang, Shuzhe and Leroy, Vincent and Cabon, Yohann and Chidlovskii, Boris and Revaud, Jerome},
  booktitle={Conference on Computer Vision and Pattern Recognition},
  year={2024}
}

@article{ObjectNavRevisited,
	author = {Dhruv Batra and Aaron Gokaslan and Ani Kembhavi and Oleksandr Maksymets and Roozbeh Mottaghi and Manolis Savva and 
	Alexander Toshev and Erik Wijmans},
    title = {{ObjectNav Revisited: On Evaluation of Embodied Agents Navigating to Objects}},
    Journal = {CoRR}, 
    volume = {2006.13171},
    year = {2020}}

\end{document}